\documentclass{article}

\PassOptionsToPackage{numbers, compress}{natbib}

\usepackage[final, nonatbib]{neurips_2021}
\usepackage[utf8]{inputenc} 
\usepackage[T1]{fontenc}    
\usepackage[pagebackref=true,breaklinks=true,colorlinks,bookmarks=false]{hyperref}
\usepackage{url}            
\usepackage{booktabs}       
\usepackage{amsfonts}       
\usepackage{nicefrac}       
\usepackage{microtype}      
\usepackage{xcolor}         
\usepackage{multirow}
\usepackage{amsmath}
\usepackage{graphicx}
\usepackage{gensymb}
\newcommand{\rulesep}{\unskip\ \vrule\ }
\usepackage{colortbl}

\title{Flood Segmentation on Sentinel-1 SAR Imagery with Semi-Supervised Learning}

\author{%
  Sayak Paul\thanks{Equal contribution. Sayak focused on the implementations and contributed to the writing, Siddha focused on ideation, experimentation, and writing.} \\
  Carted\\
  \And
  Siddha Ganju\footnotemark[1] \\
  NVIDIA, Frontier Development Lab, SpaceML\\
}

\begin{document}

\maketitle
 
\begin{abstract}
Floods wreak havoc throughout the world, causing billions of dollars in damages, and uprooting communities, ecosystems and economies. The NASA Impact Flood Detection competition tasked participants with predicting flooded pixels after training with synthetic aperture radar (SAR) images in a supervised setting. We propose a semi-supervised learning pseudo-labeling scheme that derives confidence estimates from U-Net ensembles, progressively improving accuracy. Concretely, we use a cyclical approach involving multiple stages (1) training an ensemble model of multiple U-Net architectures with the provided high confidence hand-labeled data and, generated pseudo labels or low confidence labels on the entire unlabeled test dataset, and then, (2) filter out quality generated labels and, (3) combine the generated labels with the previously available high confidence hand-labeled dataset. This assimilated dataset is used for the next round of training ensemble models and the cyclical process is repeated until the performance improvement plateaus. We post process our results with Conditional Random Fields. Our approach sets a new state-of-the-art on the Sentinel-1 dataset 
with 0.7654 IoU, an impressive improvement over the 0.60 IoU baseline. Our method, which we release with all the code and models\footnote{Our code and models are available on GitHub: \textcolor{blue}{\href{https://git.io/JW3P8}{https://git.io/JW3P8}}.}, can also be used as an open science benchmark for the Sentinel-1 dataset. 
\end{abstract}

\section{Introduction}
Flooding events are on the rise due to climate change \cite{inc-floods}, increasing sea levels and increasing extreme weather events, resulting in 10 billion dollars worth of damages annually. Scientists and decision makers can use live Earth observations data via satellites like Sentinel-1 to develop real-time response and mitigation tactics and understand flooding events. The Emerging Techniques in Computational Intelligence (ETCI) 2021 competition on Flood Detection provides SAR Sentinel-1 imagery with labeled pixels for a geographic area prior and post flood event.  SAR satellites, such as Sentinel-1, see through clouds and at night so can be used everywhere all the time, making its solutions deployable at scale. Participants are tasked with a semantic segmentation task that identifies pixels that are flooded and is evaluated with  Intersection over Union metric (IoU).

Improved flood segmentation in real-time results in delineating open water flood areas. Identifying flood levels aids in effective disaster response and mitigation. Combining the flood extent mapping with local topography generates a plan of action with downstream results including predicting the direction of flow of water, redirecting flood waters, organizing resources for distribution etc. Such a system can also recommend a path of least flood levels in real-time that disaster response professionals can potentially adopt.

Feature based machine learning techniques are also prominent \cite{mosavi} but are intractable as human annotators and featurizers cannot scale. Manual annotation in real time can easily exceed \$62,500\footnote{assuming unit economics and \$15 hourly wage} daily, and a manual solution quickly becomes intractable. Motivated by prior art \cite{Zhu05, semi_zhu, semi_chapelle, pseudolabels}, we look at semi-supervised techniques, that assume predicted labels with maximally predicted probability as ground truth, and we apply it to flood segmentation. Similar to \cite{pseudolabels, cook}, we treat pseudo labels as an entropy regularizer which eventually outperforms other conventional methods with a small subset of labeled data. For real time deployment we benchmark our solution similar to \cite{mateo-garcia, automated-flood-monitoring}, both of which utilize datasets like WorldFloods or multispectral datasets to enable flood detection, but require large scale manual annotation. Contrary to both, our work with semi supervision reduces the human-in-the-loop load and allows us to take advantage of large unannotated examples in a simple manner. Evidence indicates that post processing with Conditional Random Fields (CRF) \cite{crf, deeplabcrf, crf_meets_dl, crfrnn} may yield improved performance especially for semantic segmentation, and, satellite data \cite{HAGENSIEKER2017244}. Our work follows along similar lines when we post process with CRFs for flood segmentation. 

Our contributions are: (\textbf{1}) We propose a semi-supervised learning scheme with pseudo-labeling that derives confidence estimates from U-Net ensembles. With this we establish the new state-of-the-art flood segmenter and to the best of our knowledge we believe this is the first work to try out semi-supervised learning to improve flood segmentation models. (\textbf{2}) We show that our method is scalable through psuedo labels and generalizable through varying data distributions in different geographic locations and thus is inexpensive to deploy scalably. (\textbf{3}) Additionally, we benchmark the inference pipeline and show that it can be performed in real time aiding in real time disaster mitigation efforts. Our approach also includes uncertainty estimation, allowing disaster response teams to understand its reliability and safety. In an effort to promote open science and cross-collaboration we release all our code and models.

\section{Data}
The contest dataset consists of 66k tiled images from various geographic locations. Each RGB training tile is generated VV and VH GeoTIFF files (see raw images in Appendix \ref{supp-figures}, Figure \ref{img:all_data}) obtained via Hybrid Pluggable Processing Pipeline ``hyp3'' from the Sentinel-1 C-band SAR imagery. Data also contains swath gaps (see images in Appendix \ref{supp-figures} Figure \ref{img:noise}) where less than .5\% of an image is present; such images are not used for training. The dataset is imbalanced i.e., the proportion of images with some flood region presence is lower than the images without it, so during training, we ensure each batch contains at least 50\% samples having some amount of flood region present through stratified sampling. Flooded water may change in appearance due to added debris and such data can be captured by different sensors, but for our work we do not assume a distinction. The Red River geographic area which is predominant in the test set, is primarily an agricultural hub and recently harvested fields can look similar to floods due to low backscatter in both VV and VH polarizations. Similarly Florence which comprises of the validation set has a primarily urban setting. Such varying backscatter is relevant for performance optimizations and generalizability to test imagery (see combined images in Figure \ref{img:all_data}), and thus we combine different forms of ensembling with stacking, and, test-time augmentation helping model uncertainty and making the predictions robust. Training augmentation includes horizontal flips, rotations, and elastic transformations, and, test-time augmentations are comprised of \textit{Dihedral Group D4} \cite{sym12040548}.

\section{Methodology}
\begin{figure}[ht]
  \centering
  \includegraphics[scale=0.37]{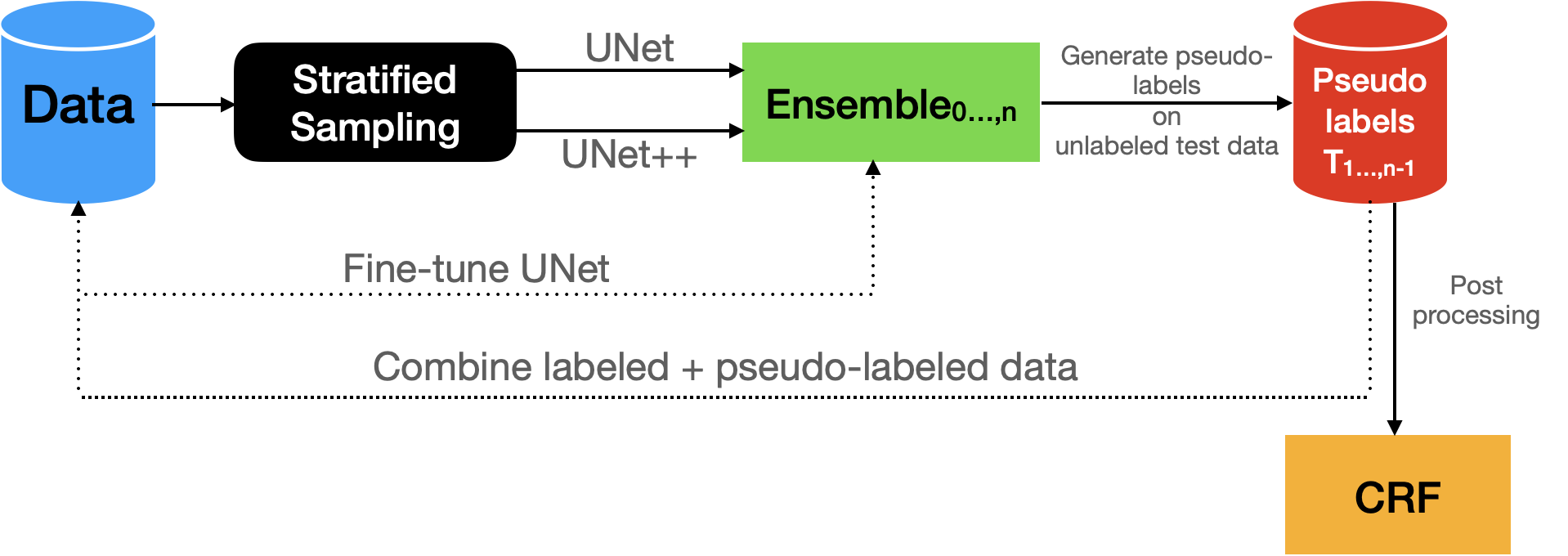}
  \caption{Our semi-supervised learning pseudo-labeling scheme that derives confidence estimates from U-Net ensembles with CRFs for post processing.}
  \label{img:pipeline}
\end{figure}

We develop a semi-supervised learning \footnote{Note: We explored another semi-supervised learning pipeline that is based on Noisy Student Training \cite{9156610}. Refer to Appendix \ref{nst} for more details.} scheme with pseudo-labeling that derives confidence estimates from U-Net ensembles motivated from \cite{NIPS2017_9ef2ed4b}. First, we train an ensemble model of multiple U-Net architectures with the provided high confidence hand-labeled data and, generated pseudo labels or low confidence labels on the entire unlabeled test dataset. We then, filter out quality generated labels and, finally, combine the quality generated labels with the previously provided high confidence hand-labeled dataset. This assimilated dataset is used for the next round of training ensemble models. This cyclical process is repeated until the performance improvement plateaus (see Figure \ref{img:pipeline}). Additionally, we post process our results with Conditional Random Fields (CRF). 

\label{strategy}

\paragraph{Step 1: Training on available data, performing inference on entire test data, and generating Pseudo Labels.}
\label{iter01}

The provided test data is only from the Red River North region which does not occur in training (Bangladesh + North Alabama + Nebraska) or validation dataset (only Florence), and thus out-of-distribution impacts were imminent. Such differences in distributions prompted us to utilize ensembling. We first train two models with U-Net \cite{unet} and U-Net++ \cite{unetpp} both with MobileNetv2 backbones \cite{mobilenetv2} and combined dice and focal loss on the available training data. Both U-Net and U-Net++ share similar inductive biases and we use them to tackle the distribution shifts, owing to different geographic locations and artifacts like recently harvested fields, urban and rural scenarios (see Appendix, Figure \ref{img:all_data}) etc. Then, create an ensemble with these two trained models.

\paragraph{Step 2: Filtering quality pseudo labels.}
\label{thresholding}
Next, we filter over the softmax output of the predictions keeping images where at least 90\% of pixels have high confidence of either flood or no flood. In the zeroth training iteration, no pseudo labels are available and training is only on the provided training dataset. For the next step, i.e., step 1 or training iteration 1, pseudo labels from iteration 0 can be used. As such training iteration n can incorporate pseudo labels from step n-1. 
For standard image classification tasks, it is common to filter the softmaxed predictions with respect to a predefined confidence threshold \cite{9156610, NEURIPS2020_06964dce, NEURIPS2020_27e9661e}. In semantic segmentation, we are extending the classification task to a per-pixel case where each pixel of an image needs to be categorized. To filter out the low-confidence predictions, we check if a pre-specified \% of the pixel values (in the range of [0, 1]) in each prediction were above a predefined threshold; mathematically denoted in (\ref{conf_eqn}). 

\begin{equation} 
\label{conf_eqn}
\begin{aligned}
\operatorname{mask}=\left(\sum_{h} \sum_{w} \tilde{Y} > c\right) > p * h * w ,
\end{aligned}
\end{equation}
where $\tilde{Y}$ is the prediction vector, $c$ and $p$ denote the confidence and pixel proportion thresholds respectively (both being 0.9 in our case), and $h$ and $w$ are the spatial resolutions of the predicted segmented maps. We compute $\tilde{Y}$ using (\ref{pred_eqn}), where $Z$ is the logit vector.

\begin{equation} 
\label{pred_eqn}
\begin{aligned}
\tilde{Y}=max \big (softmax(Z,dim=1),dim=1 \big) ,
\end{aligned}
\end{equation}
 
\paragraph{Step 3: Combining Pseudo Labels + Original Training data.}
\label{iter02}

Now, the filtered pseudo labels from the previous stage are incorporated into the training dataset. Thus, a new training dataset is created which is composed of (1) original training data with available ground truth, referred to as ``high confidence'' labels, and, (2)  filtered pseudo labels or ``low confidence'' labels on the unlabeled test dataset. This assimilated dataset is used for the next round of training individual U-Net, U-Net++ and the ensemble models. 

\paragraph{Repeat Steps 1,2,3 and Post processing with CRFs.}
With the training data now composed of the original training dataset and pseudo labels from the test dataset, we perform training from scratch of the U-Net and U-Net++ models, and fine-tuning of the U-Net from the previous iteration. Training and fine-tuning are all on the same dataset of original training data and pseudo-labeled test data. Note that the ensemble models are only used to generate predictions, and not for fine-tuning. Now, with three trained models, as before, averaged predictions are generated, and filtered to create the new set of ``weak labels''. All the data for training U-Net, U-Net++ and the fine-tuned U-Net is processed through stratified sampling as before. This cyclical process (steps 1,2,3..) are repeated until the performance improvement plateaus, about 20 epochs, post which we perform additional processing with CRFs. Ultimately for each pixel we predict as flooded or not flooded, we also produce confidence intervals that may help disaster response teams understand reliability and safety.

\section{Results}

\begin{figure}[h]
  \centering
  \includegraphics[width=.3\textwidth]{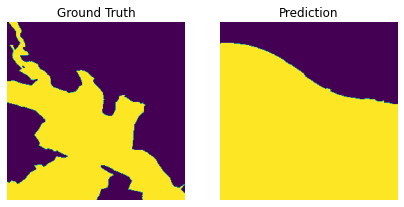} 
  \rulesep
  \includegraphics[width=.3\textwidth]{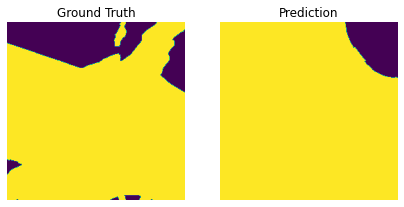}
  \caption{Random selection of Ground Truth comparison against our predictions on the hold-out test set on which the final leaderboard results are presented}
  \label{img:results}
\end{figure}

We report all the results obtained from the various approaches in Table \ref{tab:results}, and, compare against a random (all zeros to indicate non-flooded pixels, as majority pixels are not flooded) and competition provided baseline (combination of FPN \cite{fpn} with U-Net). We also provide a few random ground truth comparisons to our predictions in Figure \ref{img:results}. Since the dataset is skewed i.e., majority of pixels are not flooded, we report the IoU for the flooded pixels only. On performing test inference with the U-Net, U-Net++, the ensemble model, and averaged predictions from the ensemble, we note a performance improvement of 2-3\% IoU on average for the averaged predictions, in each step. Test-time augmentations on the test data improves our IoU by 5\% and further reduces uncertainty. Our results are uniform across all data distribution drifts available in the dataset and initial benchmarks shows that the segmentation masks are generated in approximately 3 seconds for a Sentinel-1 tile that covers an area of approximately 63,152 squared kilometers, larger than the area covered by Lake Huron, the second largest fresh water Great Lake of North America. Our work suggests CRFs are a crucial element for post processing of the predictions as they provide substantial performance improvements.

\begin{table}[!ht]
\centering
\caption{Leaderboard results for the hold-out test set. Higher IOU ($\uparrow$) is better. 
}
\label{tab:results}
\begin{tabular}{|c|c|}
\hline
\textbf{Method Description}                                                                         & \textbf{IoU $\uparrow$} \\ \hline

Random Baseline (all zeroes)                                                                                           & 0.00        \\ \hline
Competition Provided Baseline                                                                                            & 0.60         \\ \hline
Standard U-Net                                                                                            & 0.57         \\ \hline
\begin{tabular}[c]{@{}c@{}}Ensemble with CRF post processing\end{tabular}         & 0.68         \\ \hline
\begin{tabular}[c]{@{}c@{}}Pseudo labeling + Ensembles with CRF post processing \end{tabular} & \textbf{0.7654}       \\ \hline
\end{tabular}
\end{table}

\section*{Conclusion}
We recognize that it is a privilege to take part in the interdisciplinary research to reduce the impact of flooding events and as such, scaling solutions to deployment is a huge component. We developed a semi-supervised learning pseudo-labeling scheme that derives confidence estimates from U-Net ensembles. We train an ensemble model of multiple U-Net architectures with the provided high confidence hand-labeled data and, generated pseudo labels or low confidence labels on the entire unlabeled test dataset, then, filter out quality generated labels and, finally, combine the quality generated labels with the previously provided high confidence hand-labeled dataset, and, post process our results with CRFs. We show that our method can enable scalable training with data distribution drifts. 
Additionally, lack of annotated data and scaling while maintaining quality of results is also imperative. Hence our future work involves collaborating with the competition organizers and the UNOSAT team to benchmark real time runtimes and to evaluate the scalability of our solution.

\section*{Acknowledgement}
We would would like to thank the NASA Earth Science Data Systems Program, NASA Digital Transformation AI/ML thrust, and IEEE GRSS for organizing the ETCI competition
We are grateful to the Google Developers Experts program\footnote{\textcolor{blue}{\href{https://developers.google.com/programs/experts/}{https://developers.google.com/programs/experts/}}} (especially Soonson Kwon and Karl Weinmeister) for providing Google Cloud Platform credits to support our experiments and would like to thank Charmi Chokshi and domain experts 
Shubhankar Gahlot, May Casterline, Ron Hagensieker, Lucas Kruitwagen, 
Aranildo Rodrigues, Bertrand Le Saux, Sam Budd, Nick Leach, and, Veda Sunkara for insightful discussions.

\bibliographystyle{ieee_fullname}
\bibliography{references}

\clearpage
\appendix

\section*{Appendix}

\section{Implementation Details}
\label{impl-details}
Our semi-supervised learning pseudo-labeling scheme that derives confidence estimates from U-Net ensembles involved experimenting with U-Net-inspired architectures \cite{unet, unetpp} and backbones, multiple loss functions like focal loss and dice loss, combining both loss functions into one, and various training and test-time augmentation techniques.
We set seeds at various framework and package levels to enable reproducibility, and present results as the average of multiple training runs. For additional experimental configurations, we refer the readers to our code repository on GitHub\footnote{\textcolor{blue}{\href{https://git.io/JW3P8}{https://git.io/JW3P8}}}.

\paragraph{Sampling.} There is an imbalance problem in the training dataset i.e. the number of satellite images that have presence of flood regions is smaller than the number of images that do not have contain flood regions. This is why we follow a stratified sampling strategy during data loading to ensure half of the images in any given batch always contain some flood regions. Empirically we found out that this sampling significantly helped in convergence. Using this sampling strategy in our setup was motivated from a solution that won a prior Kaggle competition \cite{dsmlkz}. Research in the domain of reducing the impact of swath gaps \cite{cao2020swathgaps} also exists, but due to limited time we will explore this in future works.

\paragraph{Encoder backbone.} Throughout this work we stick to using MobileNetV2 as the encoder backbone owing to its use of pointwise convolutions which turns out to be a good fit for the problem. Since the boundary details present inside the Sentinel-1 imagery are extremely fine, point-wise convolutions are a great fit for this. Empirically, we experimented with a number of different backbones but none of their performance consistency was comparable to MobileNetV2 backbone. 

\paragraph{Segmentation architecture.} We use U-Net \cite{unet} and U-Net++ \cite{unetpp}. As before, prioritization to pointwise convolutions still stands. We avoid using architectures where dilated convolutions are used such as the DeepLab family of architectures \cite{10.1007/978-3-030-01234-2_49}. The other architectures that we tried include LinkNet~\cite{Chaurasia_2017} and MANet~\cite{zhao2020manet} but they did not produce good results. The impact of using a U-Net-based architecture with a MobileNetV2 encoder backend is emperically reported in Table \ref{tab:arch-comp}. We conclude that using this combination of U-Net-based architecture with a MobileNetV2 encoder backend for data with extremely fine segments is effective.

\begin{table}[ht]
\centering
\caption{Comparing the impact of various combinations of model architectures and encoder backbones. Using a U-Net with MobileNetV2 encoder backend outperforms amongst all others, under the same training configurations. }
\label{tab:arch-comp}
\begin{tabular}{|c|c|}
\hline
\textbf{Model Architecture + Encoder Backbone}                                                  & \textbf{IoU}  \\ \hline
U-Net + ResNet34 \cite{7780459}                                                        & 0.55          \\ \hline
U-Net + RegNetY-002 \cite{9156494}                                                    & 0.56          \\ \hline
\begin{tabular}[c]{@{}c@{}}DeepLabV3Plus +  MobileNetV2\end{tabular} & 0.52          \\ \hline
\begin{tabular}[c]{@{}c@{}}DeepLabV3Plus + RegNetY-002\end{tabular} & 0.46          \\ \hline
U-Net + MobileNetV2                                                     & \textbf{0.57} \\ \hline
\end{tabular}
\end{table}

\paragraph{Loss function.} Dice coefficient was introduced for medical workflows ~\cite{DBLP:journals/corr/MilletariNA16} to primarily deal with data imbalance. Flood imagery similar to organ or medical voxel segmentation has a large amount of imbalance with only a few pixels per image being identified as flooded. Focal loss~\cite{focalloss} assigns weight to the limited number of positive examples (flooded pixels in our case) while preventing the majority of non-flooded pixels from overwhelming the segmentation pipeline during training. We empirically noted slight improvement while using Dice loss compared to Focal loss and the two combined. 

In the following sections, we discuss the baselines and subsequent modifications we explored. To validate our approaches, we report the IoU scores on the test set obtained from the competition leaderboard (Table \ref{tab:results}).

\paragraph{Baseline model.} U-Net with a MobileNetV2 backbone and test-time augmentation. No pseudo-labeling or Conditional Random Fields (CRFs) for post processing are utilized. This gets to an IoU of 0.57 on the test set leaderboard. For the initial training of U-Net and U-Net++ (as per Section \ref{iter01}), we use Adam \cite{DBLP:journals/corr/KingmaB14} as the optimizer  with a learning rate (LR) of 1e-3\footnote{Rest of the hyperparameters were kept to their defaults as provided in \texttt{torch.optim.Adam}.} and we train both the networks for 15 epochs with a batch size of 384. For the second round of training with the initial training set and the generated pseudo-labeled dataset (as per Section \ref{iter02}, we keep all the settings similar except for the number of epochs and LR scheduling. We train the networks for 20 epochs (with the same batch size of 384) in this round to account for the larger dataset and also use a cosine decay LR schedule \cite{DBLP:conf/iclr/LoshchilovH17} since we are fine-tuning the pre-trained weights. We do not make use of weight decay for any of these stages. For additional details on the other hyperparameters, we refer the readers to our code repository on GitHub\footnote{\textcolor{blue}{\href{https://git.io/JW3P8}{https://git.io/JW3P8}}}.

\paragraph{Modified Architecture.}With the exact same configuration as the baseline, we trained a U-Net++ and got an IoU of 0.56.

\paragraph{Ensembling.} An ensemble of the baseline U-Net and U-Net++ produced a boost in the performance with 0.59 IoU. We follow a stacking-based ensembling approach where after deriving predictions from each of the ensemble members we simply taken an average of those predictions. 

We induce an additional form of ensembling with test-time augmentation. Utilizing test-time augmentation during inference was motivated due to data distribution differences, and, to better model the uncertainties and empirically we emphasize its impact in Table \ref{tab:tta}.

\begin{table}[ht]
\centering
\caption{Using test-time augmentation (TTA) during inference in our case significantly helped boost performance. The trained model in both cases is consistent with a U-Net architecture with MobileNetV2 backend.}
\label{tab:tta}
\begin{tabular}{|c|c|}
\hline
\textbf{Method}                                                        & \textbf{IoU}  \\ \hline
\begin{tabular}[c]{@{}c@{}}U-Net \end{tabular} & 0.52          \\ \hline
\begin{tabular}[c]{@{}c@{}}U-Net + TTA\end{tabular}    & \textbf{0.57} \\ \hline
\end{tabular}
\end{table}

\paragraph{Code.} Our code is in PyTorch 1.9 \cite{NEURIPS2019_9015}. We use a number of open-source packages to develop our training and inference workflows. Here we enlist all the major ones. For data augmentation, we use the \texttt{albumentations} package \cite{info11020125}. \texttt{segmentation\_models\_pytorch} (\texttt{smp} for short) package \cite{Yakubovskiy:2019} is used for developing the segmentation models. The \texttt{timm} package \cite{rw2019timm} allowed us to rapidly experiment with different encoder backbones in \texttt{smp}. Test-time augmentation during inference is performed using the \texttt{ttach} \cite{ttach} package and provides an improvement of approximately 5\%. For post processing the initial predictions, we apply CRFs leveraging the \texttt{pydensecrf} package \cite{pydensecrf}. To further accelerate the post processing time, we use the Ray framework \cite{10.5555/3291168.3291210} to achieve parallelism in applying CRF to the individual predictions. Our hardware setup includes four NVIDIA Tesla V100 GPUs. By utilizing mixed-precision training \cite{micikevicius2018mixed} (via \texttt{torch.cuda.amp}) and a distributed training setup (via \texttt{torch.nn.parallelDistributedDataParallel}) we obtain significant boosts in the overall model training time.

\section{Experiments with Noisy Student Training} 
\label{nst}
\begin{figure}[h]
  \centering
  \includegraphics[scale=0.28]{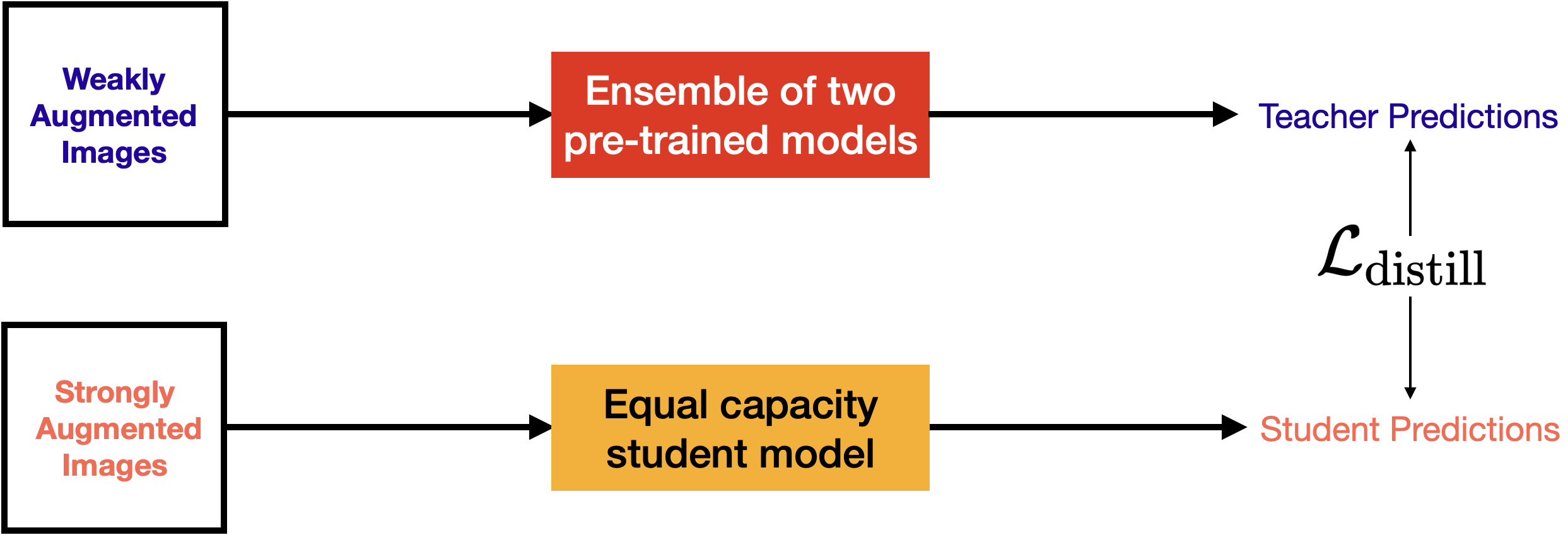}
  \caption{Our semi-supervised training pipeline based on Noisy Student Training \cite{9156610}.}
  \label{img:nst}
\end{figure}

In an effort to unify our iterative training procedure, we also experimented with techniques like the Noisy Student Training \cite{9156610} method to, but this method did not fare well. Following the recipes of \cite{9156610}, we performed self-training with noise injected only to the training data\footnote{In Noisy Student Training, noise is injected to the models as well in the form of Stochastic Depth \cite{10.1007/978-3-319-46493-0_39} and Dropout \cite{10.5555/2627435.2670313}.}. We used the ensemble of the U-Net and U-Net++ models as a teacher and a U-Net model (with MobileNetV2 backend) as a student. During training the student model our data consists of both the training and test data. This training pipeline is depicted in Figure \ref{img:nst}. With this pipeline we obtained an IoU of 0.75, which is inferior to the approach we ultimately followed. We also note that this method requires significantly less compute compared to the approach we ultimately settled with. So, if IoU can be traded with limited compute requirements, this method still yields competitive results. In regard to Figure \ref{img:nst}, $\mathcal{L}_{\text {distill}}$ is defined as per (\ref{distill_eqn}).

\begin{equation} 
\label{distill_eqn}
\begin{aligned}
\left.\mathcal{L}_{\text{distill }}=(1-\alpha) \mathcal{L}_{\text {DICE}}\left(\text {s\_preds}\left(Z_{s}\right), y\right)+\alpha \mathcal{L}_{\mathrm{KL}}(\text {(s\_preds} / \tau),(\text {t\_preds} / \tau)\right),
\end{aligned}
\end{equation}
where $s\_preds$ and $t\_preds$ denote predictions from the student and teacher networks respectively, $y$ is a vector of containing the ground-truth segmentation maps, $\alpha$ is a scalar that controls the contributions from $\mathcal{L}_{\text {DICE}}$ and $\mathcal{L}_{\mathrm{KL}}$ (KL-Divergence), and $\tau$ is a scalar denoting the temperature \cite{44873}. Note that for computing $\mathcal{L}_{\text {DICE}}$ in \ref{distill_eqn}, we use the predictions obtained from strongly augmented original training set and their ground-truth segmentation maps. 

From our experiments, we believe that with additional tweaks inspired from \cite{zou2021pseudoseg, berthelot2021adamatch} it is possible to further push this performance and we aim to explore this as future work.

\section{Supplemental Figures}
\label{supp-figures}

\begin{figure}[h]
  \centering
  \includegraphics[width=\textwidth]{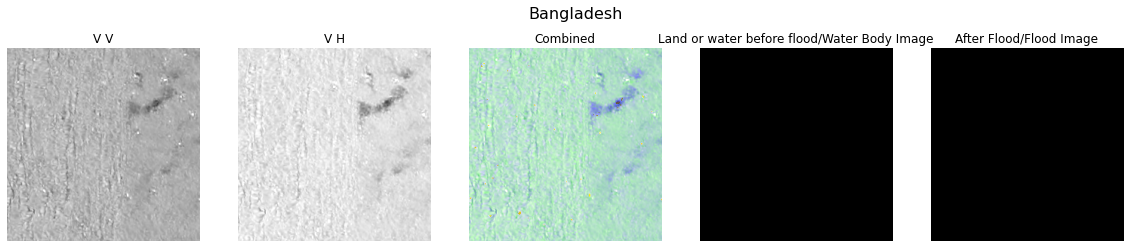}
  \includegraphics[width=\textwidth]{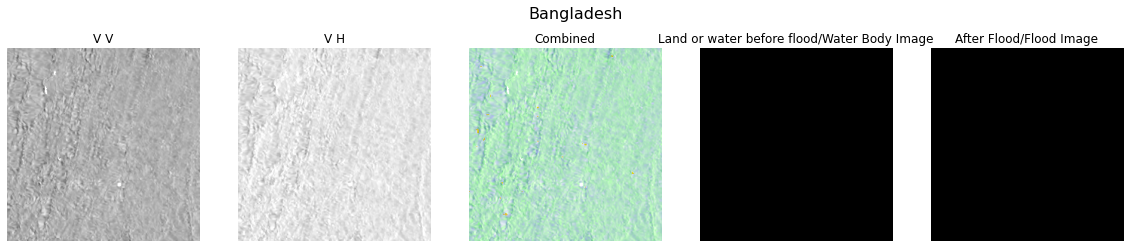}
  \includegraphics[width=\textwidth]{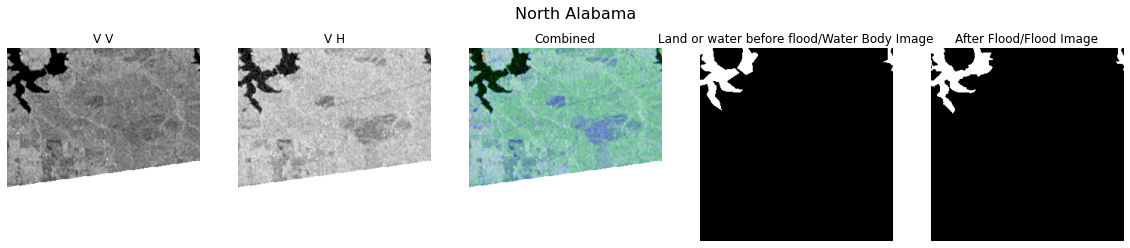}
  \includegraphics[width=\textwidth]{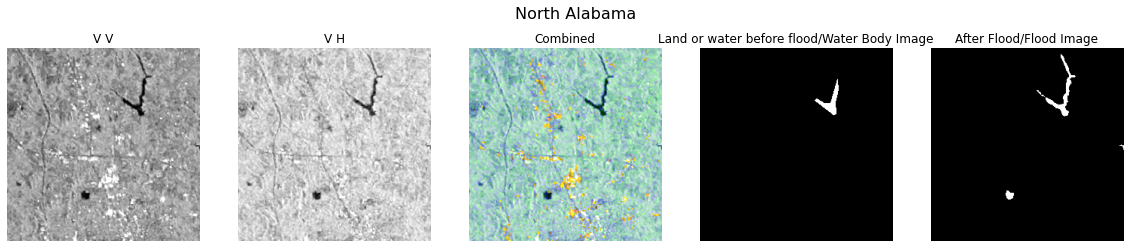}
  \includegraphics[width=\textwidth]{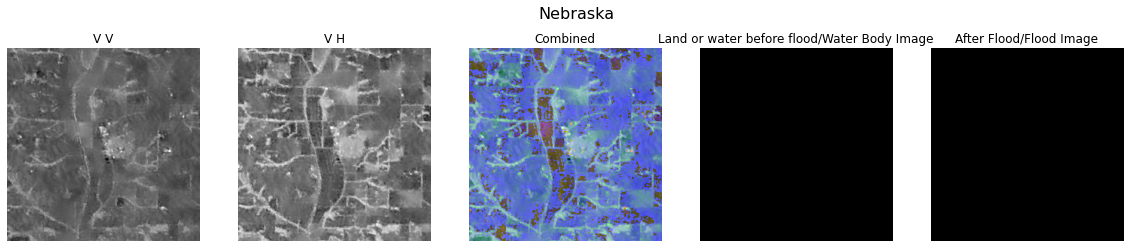}
 \includegraphics[width=\textwidth]{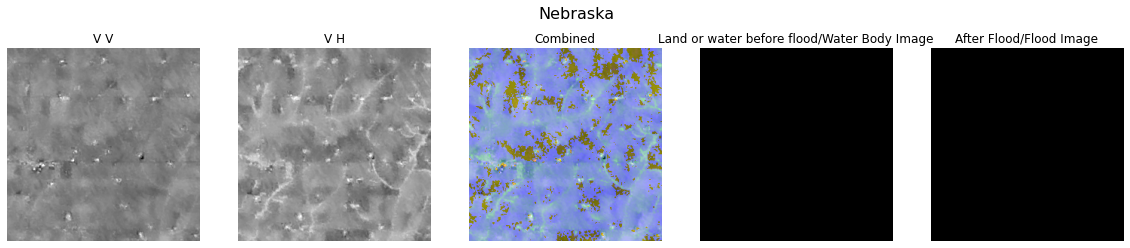}
  \caption{Raw data (VV and VH tiles) provided by the competition, and the state of water body before and after flood. The composite RGB or colour image which we use for training, is generated using the ESA Polarimetry guidelines with VV channel for red, VH channel for green and the ratio $|VV| / |VH|$ for blue. All images are a random sample from the training set. We note the visible grains in different directions potentially due to recently harvested agricultural fields from Bangladesh. 
  North Alabama show various artifacts including potential swath gaps due to differences in satellite coverage, while the RGB color range in Nebraska is unique. 
  The North Alabama image with swath gaps is kept because of at least some positive ground truth artifacts (after flood) available.}
  \label{img:all_data}
\end{figure}

\begin{figure}[h]
  \centering
  \includegraphics[width=\textwidth]{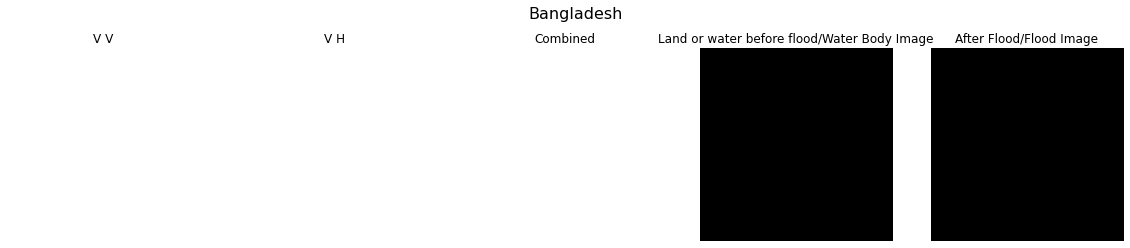}
  \includegraphics[width=\textwidth]{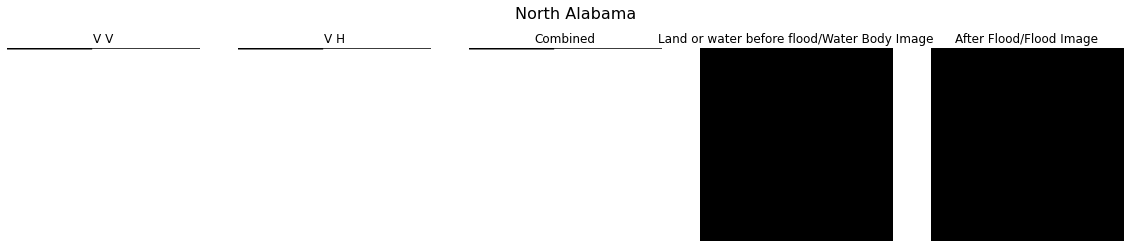}
  \includegraphics[width=\textwidth]{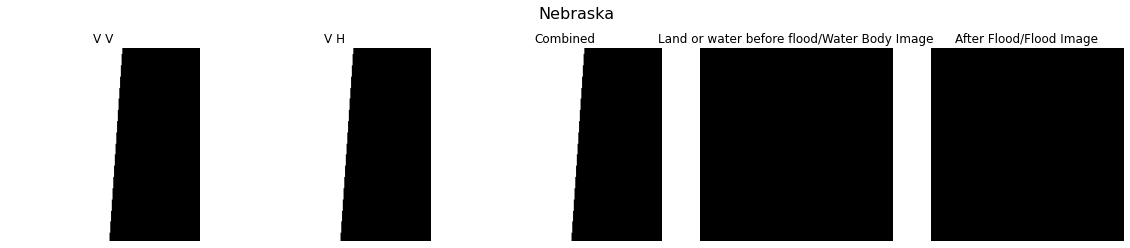}
  \caption{Noisy images either due to swath gaps or completely empty ones which occur when the VV and VH images do not align and are filtered out. Note that ground truth artifacts are unfavorable as they do not provide a positive example.}
  \label{img:noise}
\end{figure}

\end{document}